\documentclass{article}

\usepackage{scml2025}

\usepackage{multicol}
\usepackage[utf8]{inputenc} 
\usepackage[T1]{fontenc}    
\usepackage{hyperref}       
\usepackage{url}            
\usepackage{booktabs}       
\usepackage{amsfonts}       
\usepackage{amsmath}
\usepackage{amsthm}
\usepackage{amssymb}
\usepackage{amsfonts}
\usepackage{graphicx}
\usepackage{caption}
\usepackage{subcaption}
\usepackage{mathrsfs}
\usepackage{xcolor}     
\usepackage{enumitem}
\usepackage{nicefrac}       
\usepackage{microtype}      
\usepackage{cleveref}       
\usepackage{lipsum}         
\usepackage{multirow}
\usepackage{graphicx}
\usepackage{natbib}
\usepackage{doi}
\usepackage{wrapfig}
\usepackage{float}
\usepackage{mathtools}
\usepackage{bm} 

\newtheorem{theorem}{Theorem}[section]

\newtheorem{proposition}[theorem]{Proposition}
\theoremstyle{definition}
\newtheorem{definition}{Definition}[section]
\theoremstyle{remark}

\newtheorem*{example}{Example}

\title{Generalized Lie Symmetries \\in Physics-Informed Neural Operators}
\renewcommand{\shorttitle}{Generalized Lie Symmetries in PINOs}

\date{}

\newif\ifuniqueAffiliation
\uniqueAffiliationtrue

\ifuniqueAffiliation 
\author{%
	Amy Xiang Wang\thanks{Equal Contribution} \\
	New York University\\
	\texttt{xw914@nyu.edu} \\
    \And
	Zakhar Shumaylov\footnotemark[1] \\
	University of Cambridge\\
	\texttt{zs334@cam.ac.uk} \\
    \And
	Peter Zaika \\
	University of Cambridge\\
	\texttt{paz24@cam.ac.uk} \\
    \AND
    Ferdia Sherry \\
	University of Cambridge\\
	\texttt{fs436@cam.ac.uk} \\
    \And
	Carola-Bibiane Sch\"onlieb \\
	University of Cambridge\\
	\texttt{cbs31@cam.ac.uk} \\
}
\else
\usepackage{authblk}

\author[1]{%
	Xiang Wang\thanks{\texttt{xw914@nyu.edu}}}%
\author[2]{%
	Zakhar Shumaylov\thanks{\texttt{zs334@cam.ac.uk}}}%
\author[2]{Peter Zaika}
\author[2]{Ferdia Sherry}
\author[2]{Carola-Bibiane Sch\"onlieb}
\affil[1]{New York University}
\affil[2]{University of Cambridge}
\fi

\hypersetup{
pdftitle={A template for the arxiv style},
pdfsubject={q-bio.NC, q-bio.QM},
pdfauthor={David S.~Hippocampus, Elias D.~Striatum},
pdfkeywords={First keyword, Second keyword, More},
}

\begin{document}
\maketitle
\ifuniqueAffiliation \vspace{-3em}
\else \vspace{-3em}
\fi
\begin{abstract}
Physics-informed neural operators (PINOs) have emerged as powerful tools for learning solution operators of partial differential equations (PDEs). Recent research has demonstrated that incorporating Lie point symmetry information can significantly enhance the training efficiency of PINOs, primarily through techniques like data, architecture, and loss augmentation. In this work, we focus on the latter, highlighting that point symmetries oftentimes result in no training signal, limiting their effectiveness in many problems. 
To address this, we propose a novel loss augmentation strategy that leverages evolutionary representatives of point symmetries, a specific class of generalized symmetries of the underlying PDE. These generalized symmetries provide a richer set of generators compared to standard symmetries, leading to a more informative training signal. We demonstrate that leveraging evolutionary representatives enhances the performance of neural operators, resulting in improved data efficiency and accuracy during training.
\end{abstract}


\begin{multicols}{2}
\section{Introduction}

Deep neural networks are increasingly used for scientific computing, particularly in simulating complex physical systems governed by partial differential equations (PDEs) \cite{han2018solving},  where traditional methods struggle \cite{raissi2019physics}.

Physics-informed neural networks (PINNs) \citep{raissi2019physics} integrate physical laws and constraints directly into the learning process, demonstrating success in various applications \citep{raissi2019physics,sun2020surrogate,zhu2019physics,karumuri2020simulator,sirignano2018dgm,zhang2024biophysics}. However, classical PINNs face challenges under varying parameters and boundary conditions and often encounter training challenges due to unbalanced gradients and non-convex loss landscapes \citep{grossmann2023can,krishnapriyan2021characterizing, mullerposition}.
Neural operators offer a compelling alternative, which unlike standard feedforward neural networks, learn mappings between infinite-dimensional function spaces, directly connecting parameters and initial/boundary conditions to PDE solutions \cite{li2021physics,wang2021learning}. This eliminates the need for independent simulations and, when combined with physics-informed loss functions, often improves generalization and physical validity. 
We focus on this class of approaches.


In a separate direction, the field of \emph{geometric deep learning} has demonstrated the practical and theoretical benefits of incorporating symmetry information into neural networks, leading to improved performance through beneficial inductive biases \citep{bronstein2021geometric, brehmer2024doesequivariancematterscale}. For example, the translational equivariance of convolutional neural networks (CNNs) \citep{fukushima1982neocognitron,lecun1989backpropagation} has been suggested as an important reason for their successes in tasks such as image classification \citep{krizhevsky2012imagenet, he2016deep}.

Similarly, recent work in neural operators has shown that incorporating PDE symmetries improves generalization and training efficiency. However, most research has focused on Lie point symmetries due to their well-established theoretical foundation and systematic derivation methods \citep{olver1993applications, ibragimov1995crc, Baumann1998MathLieAP}. In contrast, the potential of generalized symmetries remains largely unexplored due to their unsuitability for data augmentation or equivariant architectures. 

This paper demonstrates that generalized PDE symmetries can nonetheless provide valuable inductive biases when incorporated through loss augmentation.

\subsection{PDE Symmetries in Deep Learning}
Incorporation of point symmetries into neural solvers has received significant attention, leading to the development of many approaches that can be broadly classified into three categories:
    \vspace{-1em}
    \paragraph{Data augmentation}is the simplest way to add symmetry information into neural PDE solvers \citep{brandstetter2022lie, 10016380} by augmenting the training data with transformed versions obtained by applying symmetry operations. While straightforward to implement, this approach can increase training costs.
    \vspace{-1em}
    \paragraph{Loss augmentation}is an alternative way to inject symmetry information by appropriately regularizing the loss using symmetry information \citep{akhoundsadegh2023lie,li2023utilizing,zhang2023enforcing}. This often leads to improved generalization and sample efficiency, although it may not guarantee equivariant models.
    \vspace{-1em}
    \paragraph{Architecture augmentation}aims to directly embed symmetries into the model architecture. Given the complexity of point symmetry groups associated with PDEs, this requires either general techniques \citep{shumaylov2024lie} or exploiting simpler subgroups with existing architectures \citep{wang2021incorporating}. 
    
Beyond these, alternative approaches exist, based on problem reformulations \citep{arora2024invariant, lagrave2022equivariant} or those focused on other types of structures, e.g. from Hamiltonian dynamics \citep{canizares2024hamiltonian,canizares2024symplectic,tanaka2024neuraloperatorsmeetenergybased}.
\section{PDE Symmetries}\label{sec:background}
Here, we introduce the notion of PDE symmetries. The formal presentation here closely follows \citep{olver1993applications}, and the reader is strongly encouraged to read it for more in-depth discussions of the concepts. 

In simple words, PDE point symmetries are transformations that map solutions of a given PDE to other solutions of the same PDE. To derive these symmetries, traditionally the focus is shifted to Jet spaces, where the problem of studying PDE symmetries is transformed into the simpler task of studying symmetries of algebraic equations.  By employing one-parameter groups of transformations, it becomes possible to systematically derive all such symmetries by examining their infinitesimal actions.

Suppose we are considering a system $\Delta$ of $n$-th order differential equations involving $p$ independent  $x=\left(x^1, \dots, x^p\right)\in X$, and $q$ dependent variables $u=\left(u^1, \dots , u^q\right) \in U$. 
\paragraph{Jet Spaces and Prolongations:} 
Consider $f: X \rightarrow U$ smooth and let $U_k$ be the Euclidean space, endowed with coordinates $u_J^\alpha=\partial_J f^\alpha(x)$ in multi-index notation so as to represent the above derivatives. Furthermore, set $U^{(n)}=U \times U_1 \times \cdots \times U_n$ to be the Cartesian product space, whose coordinates represent all the derivatives of functions $u=f(x)$ of all orders from 0 to $n$. The total space $X \times U^{(n)}$, whose coordinates represent the independent variables, the dependent variables and the derivatives of the dependent variables up to order $n$ is called the $n$-th order jet space of the underlying space $X \times U$. 
Given a smooth function $u=f(x)$, there is an induced function $u^{(n)}=\operatorname{pr}^{(n)} f(x)$, called the $n$-th prolongation of $f$, defined by the equations $u_J^\alpha=\partial_J f^\alpha(x).$
Thus $\mathrm{pr}^{(n)} f$ is a function from $X$ to the space $U^{(n)}$, and for each $x$ in $X$, $\operatorname{pr}^{(n)} f(x)$ is a vector representing values of $f$ and all its derivatives up to order $n$ at the point $x$. 
The PDE is a system of equations $\Delta\left(x, u^{(n)}\right)=0,$ with $\Delta: X \times U^{(n)} \rightarrow \mathbb{R}^l$ smooth, determining a subvariety
$$
S_{\Delta}=\left\{\left(x, u^{(n)}\right): \Delta\left(x, u^{(n)}\right)=0\right\} \subset X \times U^{(n)}
$$
A symmetry group of the system $\Delta$ will be a local group of transformations, $G^{\Delta}$, acting on some open subset $M \subset X \times U$ such that ``$G$ transforms solutions of $\Delta$ to other solutions of $\Delta$'', i.e. leaving $S_{\Delta}$ invariant.
\paragraph{Prolongations of actions}
Now suppose $G$ acts on an open $M \subset X \times U$, i.e. independent and dependent variables. There is an induced local action of $G$ on the $n$-jet space $M^{(n)}$, called the $n$-th prolongation of $G$ denoted $\mathrm{pr}^{(n)} G$. This prolongation is defined so that it transforms the derivatives of functions $u=f(x)$ into the corresponding derivatives of the transformed function $\tilde{u}=\tilde{f}(\tilde{x})$. 
Consider now $v$ a vector field on $M$, with a corresponding one-parameter group $\exp (\varepsilon v)$. The $n$-th prolongation of ${v}$ is denoted $\operatorname{pr}^{(n)} v$, will be a vector field on the $n$-jet space $M^{(n)}$, and is defined to be the infinitesimal generator of the corresponding prolonged one-parameter group $\operatorname{pr}^{(n)}[\exp (\varepsilon v)]$. Relevance of these concepts is highlighted by the following theorem:
\begin{theorem}[2.31 in \citep{olver1993applications}]\label{thm:pr_theorem}
Suppose $\Delta\left(x, u^{(n)}\right)=0$ is a system of differential equations of maximal rank defined over $M \subset X \times U$. If $G$ is a local group of transformations acting on $M$, and
\[
\operatorname{pr}^{(n)} {v}\left[\Delta\left(x, u^{(n)}\right)\right]=0, \;\; \text{whenever} \;\; \Delta\left(x, u^{(n)}\right)=0
\]
for every infinitesimal generator ${v}$ of $G$, then $G$ is a symmetry group of the system.  
\end{theorem}
\begin{figure*}[t]
    \begin{minipage}{0.22\textwidth}
        \centering
        \includegraphics[width=\textwidth]{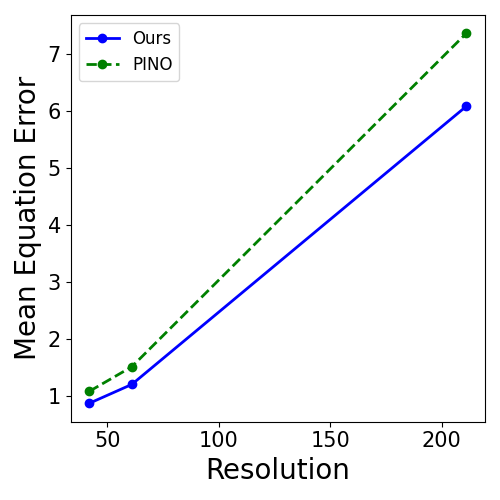}
    \end{minipage}
    \begin{minipage}{0.22\textwidth} 
        \includegraphics[width=\textwidth]{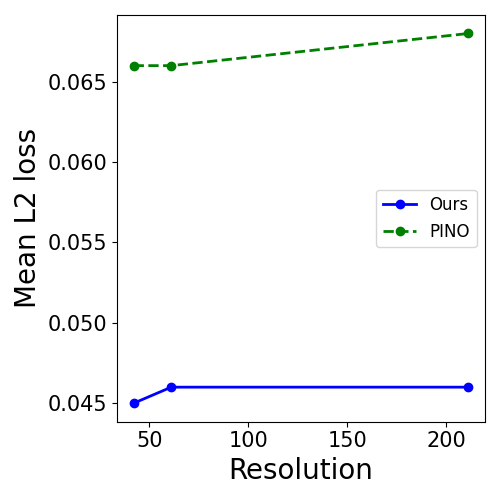} 
    \end{minipage}
    \begin{minipage}{0.22\textwidth} 
        \includegraphics[width=\textwidth]{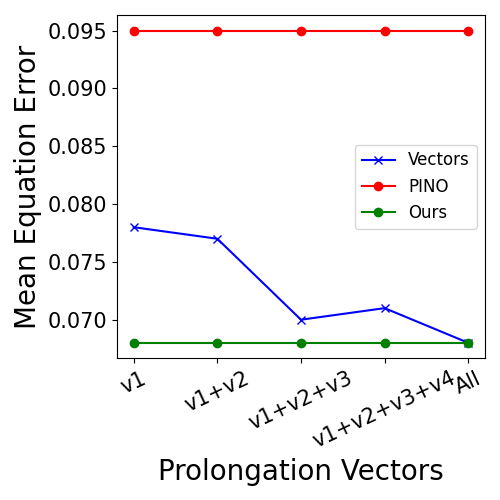} 
    \end{minipage}
    \begin{minipage}{0.30\textwidth}  
	\includegraphics[width=\textwidth]{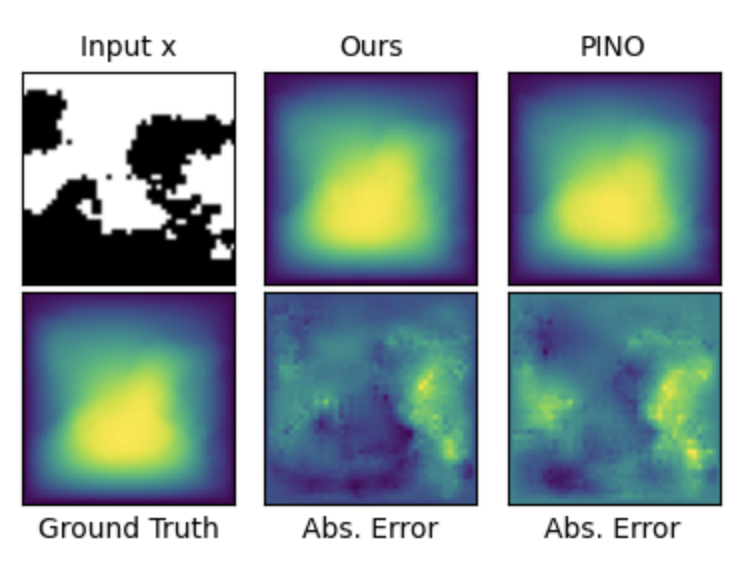}
    \end{minipage}
    \caption{\textbf{Left} (first two): Darcy Flow zero-shot evaluation by resolution. \textbf{Middle}: Burgers' equation symmetry regularization effectiveness by vector. \textbf{Right}: Darcy Flow prediction results comparison with 100 training samples}
    \label{fig:burger}
\end{figure*}
\paragraph{Generalized Symmetries}
We can consider some vector field acting on $M$ of the form
$$
v=\sum_{i=1}^p \xi^i(x, u) \frac{\partial}{\partial x^i}+\sum_{\alpha=1}^q \phi_\alpha(x, u) \frac{\partial}{\partial u^\alpha},
$$
and provided the coefficient functions $\xi^i, \phi_\alpha$ depend only on $x$ and $u$, $v$ generates a one-parameter group of transformations $\exp (\varepsilon v)$ acting pointwise. A significant generalization of the notion of symmetry group is obtained by relaxing this geometrical assumption, and allowing the coefficients $\xi^i, \phi_\alpha$ to also depend on derivatives of $u$. 
\begin{definition}[Generalized Vector Fields]
A generalized vector field is an expression of the form
$$
v=\sum_{i=1}^p \xi^i[u] \frac{\partial}{\partial x^i}+\sum_{\alpha=1}^q \phi_\alpha[u] \frac{\partial}{\partial u^\alpha}
$$
in which $\xi^i$ and $\phi_\alpha$ are smooth differential functions.   
\end{definition}
\begin{example}
An example of a generalized vector field is 
$v=u_x \partial_u$, which admits the following prolongation:
$$
\operatorname{pr} {v}=u_x \frac{\partial}{\partial u}+u_{x x} \frac{\partial}{\partial u_x}+u_{x t} \frac{\partial}{\partial u_t}+u_{x x x} \frac{\partial}{\partial u_{x x}}+\cdots
$$
and thus turns out to be a generalized symmetry of the Burgers' \Cref{eq:burgers}, as $\operatorname{pr} {v} \left[\Delta_B\right]=0$.
\end{example}
\paragraph{Evolutionary Vector Fields}
Among all the generalized vector fields, those in which the coefficients $\xi^i[u]$ of the $\partial / \partial x^i$ are zero play a particularly important role.
\begin{definition}[5.4 in \cite{olver1993applications}]
Let $Q[u]=\left(Q_1[u], \ldots, Q_q[u]\right)$ be a $q$-tuple of differential functions. The generalized vector field
$$
v_Q=\sum_{\alpha=1}^q Q_\alpha[u] \frac{\partial}{\partial u^\alpha}
$$
is called an evolutionary vector field, and $Q$ is called its characteristic.
\end{definition}
Any generalized vector field ${v}$ has an associated evolutionary representative ${v}_Q$ with characteristic
$Q_\alpha=\phi_\alpha-\sum_{i=1}^p \xi^i \partial u^\alpha / \partial x^i$, determining essentially the same symmetry, as illustrated by the following proposition:
\begin{proposition}[5.5 in \citep{olver1993applications}]\label{prop:evolutionary}
A generalized vector field ${v}$ is a symmetry of a system of differential equations if and only if its evolutionary representative ${v}_Q$ is.
\end{proposition}
Evolutionary vector fields are particularly appealing as they capture the essence of the symmetry, discarding the dependent variable information, while also admitting much simpler expressions for their prolongations.

However, it is important to note that these are not symmetries in the traditional sense, as they no longer act on $X\times U$. For a more in-depth discussion, see \cite{olver1993applications}.

\section{Methodology}\label{sec:methodology}
\begin{table*}[t]
\centering
\caption{Comparison of $L_2$ loss and equation error for Darcy Flow and Burgers' equation with varying sample sizes, using our symmetry-based method, PINO (no symmetry), and the method from \cite{akhoundsadegh2023lie,zhang2023enforcing}.}
\renewcommand{\arraystretch}{1.0} 
\resizebox{0.8\linewidth}{!}{ 
\begin{tabular}{lllccc}
\toprule
Equation & No. Samples & Metric & Symmetry (Ours) & No Symmetry (PINO) & Symmetry \cite{akhoundsadegh2023lie,zhang2023enforcing} \\
\midrule
\multirow{8}{*}{Darcy Flow} 
& 100  & \multirow{4}{*}{$L_2$ Loss}  & $\bm{0.046 \pm 1e-3}$  & \multicolumn{2}{c}{$0.066 \pm 1e-2$} \\ 
& 250  &   & $\bm{0.029 \pm 7e-4}$  & \multicolumn{2}{c}{$0.323 \pm 8e-4$} \\ 
& 500  &   & $\bm{0.014 \pm 3e-4}$  & \multicolumn{2}{c}{$0.017 \pm 4e-4$} \\ 
& 1000 &   & $\bm{0.011 \pm 2e-3}$  & \multicolumn{2}{c}{$0.013 \pm 3e-4$} \\ 
\cmidrule{2-6}
& 100  & \multirow{4}{*}{Eqn. Error} & $\bm{1.202 \pm 1e-2}$  & \multicolumn{2}{c}{$1.551 \pm 1e-2$} \\ 
& 250  &  & $\bm{0.621 \pm 7e-3}$  & \multicolumn{2}{c}{$0.925 \pm 1e-2$} \\ 
& 500  &  & $\bm{0.579 \pm 7e-3}$  & \multicolumn{2}{c}{$0.723 \pm 9e-3$} \\ 
& 1000 &  & $\bm{0.491 \pm 6e-3}$  & \multicolumn{2}{c}{$0.583 \pm 7e-3$} \\ 
\midrule
\multirow{6}{*}{Burgers' Eqn.}
& 25   & \multirow{3}{*}{$L_2$ Loss} & $\bm{0.082 \pm 6e-3}$ & $0.089 \pm 6e-3$ & $0.099 \pm 6e-3$ \\
& 50   &  & $\bm{0.044 \pm 3e-3}$ & $\bm{0.044 \pm 3e-2}$ & $0.050 \pm 4e-3$ \\
& 100  &  & $\bm{0.018 \pm 1e-3}$ & $0.019 \pm 1e-3$ & $0.022 \pm 1e-3$ \\
\cmidrule{2-6}
& 25   & \multirow{3}{*}{Eqn. Error} & $\bm{0.182 \pm 3e-2}$ & $0.217 \pm 4e-2$ & $0.440 \pm 6e-2$ \\
& 50   &  & $\bm{0.068 \pm 1e-2}$ & $0.095 \pm 2e-2$ & $0.217 \pm 4e-2$ \\
& 100  &  & $\bm{0.023 \pm 4e-3}$ & $0.024 \pm 4e-3$ &  $0.068 \pm 1e-2$ \\
\bottomrule
\end{tabular}}
\label{tab:combined_all_vertical_concise}
\end{table*}
Based on the discussion in \Cref{sec:background}, instead of considering symmetry vector fields, we propose to use evolutionary representatives of these vector fields instead. These turn out to be more informative in practice, as illustrated below.

To understand the proposed methodology, we first recap in simpler notation, the method of \citep{akhoundsadegh2023lie,zhang2023enforcing}. Clasically, PINNs aim to solve the PDE by minimizing the residual 
\begin{equation}\label{eq:pinn_loss}
\min_\theta\; \|\Delta[u_{\theta}]\|^2_2.
\end{equation}
In \citep{akhoundsadegh2023lie,zhang2023enforcing}, it is proposed that for a Lie algebra $v_i$, it is beneficial to minimize (for some $\gamma>0$):
\begin{equation}\label{eq:sympinn_loss}
\min_\theta\; \|\Delta[u_{\theta}]\|^2_2 + \gamma \sum_i\left\|\operatorname{pr} v_i\left[\Delta\right][u_{\theta}]\right\|^2_2,    
\end{equation}
as by \Cref{thm:pr_theorem}, zero loss solutions of \Cref{eq:pinn_loss} are also zero loss solutions of \Cref{eq:sympinn_loss}. 
Based on \Cref{prop:evolutionary}, we instead propose to use the evolutionary representatives of $v_i$, via 
\begin{equation}\label{eq:g_sympinn_loss}
\min_\theta\; \|\Delta[u_{\theta}]\|^2_2 + \gamma \sum_i\left\|\operatorname{pr} \left[v_i\right]_{Q}\left[\Delta\right][u_{\theta}]\right\|^2_2,    
\end{equation}
as motivated by the following example: 
\begin{example}[Burgers]
    Consider Burgers' \Cref{eq:burgers}, and consider the symmetry generated by $v_1 = \partial_x$. Its evolutionary representative, as above, is $\left[v_1\right]_{Q} = -u_x\partial_u$.
    We can derive both prolongation actions as: 
\begin{equation}
\operatorname{pr} v_1 \left[\Delta_B\right] = 0, \quad
\operatorname{pr} \left[v_1\right]_{Q} \left[\Delta_B\right] = -D_x\left[\Delta_B\right],     
\end{equation}
where $D_x$ denotes the total $x$ derivative. This shows that standard point symmetries result in no extra terms in \Cref{eq:sympinn_loss}, unlike \Cref{eq:g_sympinn_loss}. Same occurs for all other generators for both Burgers' equation (\Cref{ap:burgers}) and Darcy Flow (\Cref{ap:darcy}).
\end{example}
\section{Experiments}
In this section we present numerical results showcasing the utility of generalized symmetries for training PINOs. This is illustrated for the 2D Darcy flow and the 1D Burgers' equation. Our code is publicly available at  \url{https://github.com/xiwang129/GPS_PINO}.
\subsection{Darcy Flow} 
We consider solving the 2D Darcy Flow equation 
\begin{equation}\label{eq:darcy}
  \Delta_D \coloneqq \nabla \cdot \left(k(x) \nabla u(x)\right) + f(x) = 0, 
\end{equation}
on the domain $x \in [0,1]^2$ with $u(x) = 0$ on the boundary, where $k(x)$ is the permeability field and $f(x)$ is the source term. We tested our approach on a $61 \times 61$ resolution dataset downsampled from a $421 \times 421$ resolution dataset and set $f({x})=1$ as in \cite{li2021physics}. We trained a 2D Fourier neural operator model for 300 epochs using the proposed generalized symmetry loss in \Cref{eq:g_sympinn_loss}. As is common with PINOs, in this example we also include a data loss term enforcing initial and boundary conditions, and a trajectory matching loss.
%
The trained model is tested on 500 samples. \Cref{tab:combined_all_vertical_concise} indicates that the proposed symmetry regularization improves prediction in terms of both $L_{2}$ and equation losses, being more sample efficient in minimizing the PDE residuals.
As shown in \Cref{eq:darcy_sym_tara}, the point symmetry method of \cite{akhoundsadegh2023lie,zhang2023enforcing} in \Cref{eq:sympinn_loss} generates no extra terms, being equivalent to a standard PINO.   



Moreover, we experimented with zero-shot prediction by testing the trained $61\times61$ model on other resolutions: $40\times40$ and $211\times211$. \Cref{fig:burger} left illustrates that the resulting model consistently outperforms the baselines in both $L_2$ and equation loss.

\subsection{Burgers' equation}
\begin{figure}[H]
	\begin{center}	\includegraphics[width=0.5\textwidth]{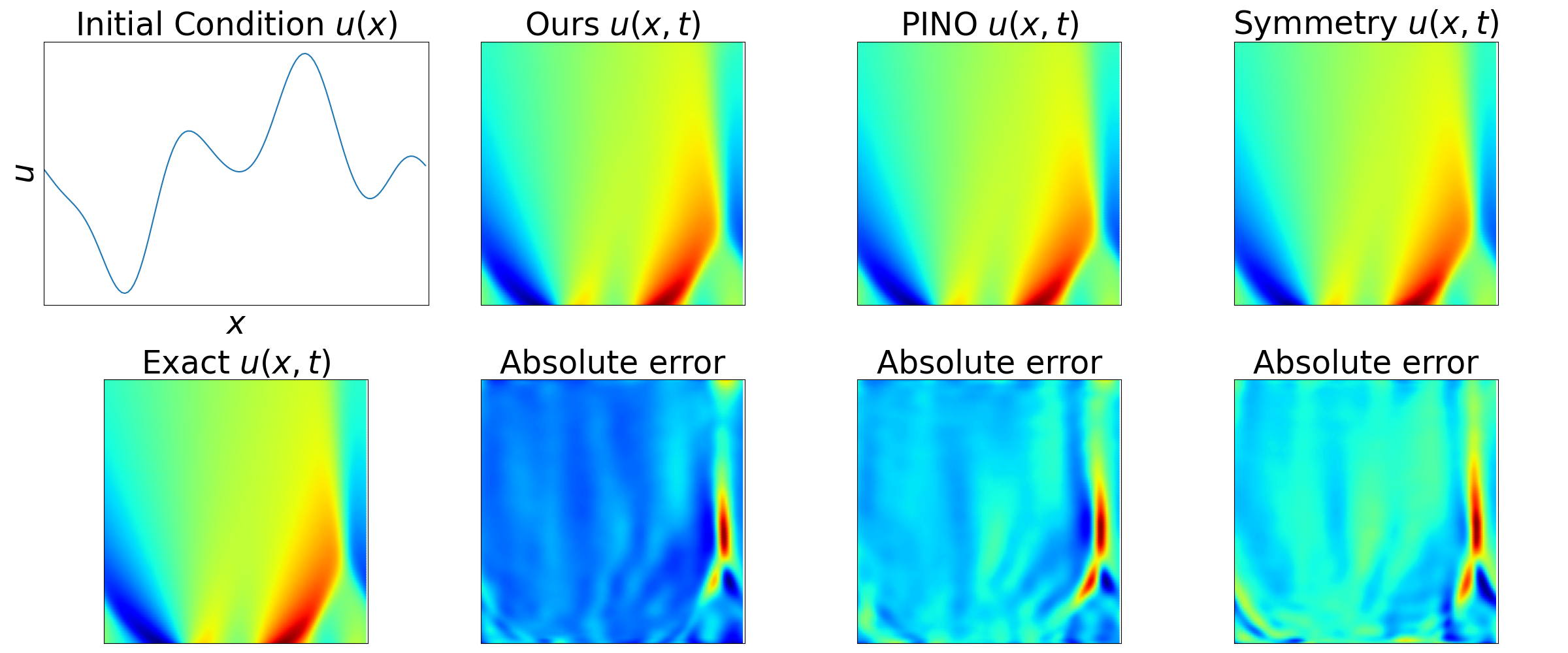}
	\end{center}
	\caption{Visual comparison of trajectory predictions for Burgers' equation trained with 50 samples.}
\label{fig:burgers_pred}
\end{figure}

We consider solving the 1D Burgers' equation 
\begin{equation}\label{eq:burgers}
  \Delta_B \coloneqq u_t + u u_x - \nu u_{xx} = 0,
\end{equation}
on the domain $x \in [0,1]$ and $t \in [0,1]$ with initial condition $u(x,0) = u_0(x)$, for which we generated the 1D Burgers' dataset following \cite{rosofsky2023applications}, using a $128 \times 100$ grid and setting the viscosity coefficient to $\nu=0.01$. We compared the proposed method in \Cref{eq:g_sympinn_loss}  with \cite{akhoundsadegh2023lie,zhang2023enforcing} defined in \Cref{eq:sympinn_loss}. \Cref{tab:combined_all_vertical_concise} presents the results, demonstrating that the proposed method consistently achieves either superior or comparable performance to the baselines with respect to both equation error and 
$L_2$ error. Following experimental details of \cite{akhoundsadegh2023lie}, the residual for this equation does not appear directly in the training objective. Figure \ref{fig:burgers_pred} provides a representative prediction showcasing the improved error achieved by our method.



We further evaluated the impact of each of the Lie algebra vectors from \Cref{eq:burgers_sym} on the resulting equation error, by adding one term at a time. An example is illustrated in the middle pane of \Cref{fig:burger}, showcasing that the introduction of even a single Lie algebra vector yields a substantial reduction in the equation error, while the benefit of incorporating additional vectors becomes marginal.


\subsection{Ablation studies}
We evaluated the robustness, generalization ability, and stability of our approach and PINO by adding various levels of noise into the training dataset using the Darcy flow equation with 500 samples, as  shown in Table \ref{tab:noise_table}. The generalization gap, calculated as the difference between the $L_2$ loss on the test and training sets, increases slightly with higher noise level. This suggests that our approach generalizes reasonably well and resists overfitting to noisy data.

For robustness, both test $L_2$ loss and the equation error show a steady increase as the noise level rises, indicating that the model remains relatively robust up to $5\%$ noise level. In comparison, our approach consistently achieves lower $L_2$ loss and the equation error than PINO as the noise level increase. 

\begin{table*}
\centering
\caption{Ablation study of model robustness, generalization, and stability using Darcy flow}
\renewcommand{\arraystretch}{1.0}
\resizebox{\linewidth}{!}{ 
\begin{tabular}{lcccccccc}
\toprule
\multirow{2}{*}{Metric} & \multicolumn{2}{c}{Clean data ($0\%$)} & \multicolumn{2}{c}{Noise ($1\%$)} & \multicolumn{2}{c}{Noise ($5\%$)} & \multicolumn{2}{c}{Noise ($10\%$)} \\
\cmidrule(lr){2-3} \cmidrule(lr){4-5} \cmidrule(lr){6-7} \cmidrule(lr){8-9}
 & Ours & PINO & Ours & PINO & Ours & PINO & Ours & PINO \\
\midrule
$L_2$ Loss (Train)  & $\bm{0.007 \pm 6e-6}$ & $0.007 \pm 6e-6$ & $0.007 \pm 6e-6$ & $0.007 \pm 6e-6$ & $0.008 \pm 6e-6$ & $0.008 \pm 6e-6$ & $0.010 \pm 4e-6$ & $0.010 \pm 5e-6$ \\
$L_2$ Loss (Test)   & $\bm{0.016 \pm 4e-4}$ & $0.019 \pm 5e-4$  & $0.018 \pm 4e-4$ & $0.019 \pm 5e-4$ & $0.019 \pm 4e-4$ & $0.020 \pm 4e-4 $ & $0.022 \pm 4e-4$ & $0.022 \pm 5e-4$  \\
Eqn. error (Train)  & $\bm{0.007 \pm 6e-4}$ & $0.007 \pm 6e-4$  & $0.092 \pm 5e-4$ & $0.091 \pm 5e-4$ & $0.264 \pm 1e-3$ & $0.268 \pm 1e-3$ & $0.461 \pm 2e-3$ & $0.488 \pm 2e-3$   \\
Eqn. error (Test)   & \bm{$0.569 \pm 7e-3$} & $0.701 \pm 9e-3$  & $0.624 \pm 8e-3$ & $0.723 \pm 9e-2$ & $0.722 \pm 8e-3$ & $0.723 \pm 9e-3$ & $0.827 \pm 4e-3$ & $0.918 \pm 9e-3$  \\
\midrule
Generalization gap  & $\bm{0.009 \pm 4e-4}$ & $0.012 \pm 5e-4$ & $0.011 \pm 4e-4$ & $0.012 \pm 5e-4$ & $0.011 \pm 4e-4$ & $0.012 \pm 4e-4$ & $0.012 \pm 4e-4$ & $0.012 \pm 5e-4$ \\
Stability gap       & $0.562 \pm 7e-3$ & $0.694 \pm 9e-3$ & $0.532 \pm 8e-3$ & $0.632 \pm 9e-3$ & $0.458 \pm 8e-3$ & $0.455 \pm 9e-3$ & $\bm{0.366 \pm 4e-3}$ & $0.430 \pm 9e-3$ \\
\bottomrule
\end{tabular}}
\label{tab:noise_table}
\end{table*}

However, noise level does impact the the model's ability to capture the underlying physics well as reflected by the stability gap, defined as the difference between the equation errors on the test and training sets. Notably,  while both the training and test equation errors increase with noise, the test equation error grows more slowly. Compared to PINO, our approach achieves a lower stability gap by having smaller equation errors, indicating that incorporating symmetry enhances its ability to learn the underlying physical laws.


\section{Conclusion}


This work demonstrates that evolutionary representatives of point symmetries enhance data efficiency in training PINOs, resulting in a useful training signal even when the underlying point symmetries themselves can not be used. 

While this study focused on point symmetries, the framework presented is readily applicable to any generalized symmetry. We hypothesize that incorporating such symmetries can similarly improve trainability and we leave a proper evaluation of this to future work.

We further posit that the observed efficiency partly arises from the inducement of a Sobolev-type norm \cite{son2023sobolev}. Crucially, our approach offers a systematic way for selecting the appropriate norm based on the governing PDE. This connection between generalized symmetries, Sobolev norms, and improved training warrants further investigation.


\section*{Acknowledgments}
CBS acknowledges support from the Philip Leverhulme Prize, UK, the Royal Society Wolfson Fellowship, UK, the EPSRC advanced career fellowship, UK EP/V029428/1, EPSRC grants EP/S026045/1 and EP/T003553/1, EP/N014588/1, EP/T017961/1, the Wellcome Innovator Awards, UK 215733/Z/19/Z and 221633/Z/20/Z, the European Union Horizon 2020 research and innovation programme under the Marie Skłodowska-Curie grant agreement No.777826 NoMADS, the Cantab Capital Institute for the Mathematics of Information, UK and the Alan Turing Institute, UK. FS acknowledges support from the EPSRC advanced career fellowship EP/V029428/1. ZS acknowledges support from the Cantab Capital Institute for the Mathematics of Information, Christs College and the Trinity Henry Barlow Scholarship scheme. AXW acknowledges the support from NYU HPC.

\bibliographystyle{plain}
\bibliography{references} 

\appendix
\newpage

\section{Burgers' Equation Symmetries}
\label{ap:burgers}

In this appendix, we provide detailed derivations of the evolutionary representatives and their prolongations for the symmetries of Burgers' equation.

Burgers' equation is given by:
\[
\Delta_B \coloneqq u_t + u u_x - \nu u_{xx} = 0
\]

The Lie algebra of symmetries for Burgers' equation includes the following generators:
\begin{align*}
& v_1 = \partial_x, \\
& v_2 = \partial_t, \\
& v_3 = 2t \partial_t + x \partial_x - u \partial_u, \\
& v_4 = t \partial_x + \partial_u, \\
& v_5 = t^2 \partial_t + t x \partial_x + (x - t u) \partial_u.
\end{align*}
It turns out to be rather simple to find evolutionary representatives of these as: 
\begin{align*}
& \left[v_1\right]_Q = -u_x \partial_u, \\
& \left[v_2\right]_Q = -u_t \partial_u, \\
& \left[v_3\right]_Q = -(u + x u_x + 2 t u_t) \partial_u, \\
& \left[v_4\right]_Q = -(t u_x - 1) \partial_u, \\
& \left[v_5\right]_Q = (x - t u - t x u_x - t^2 u_t) \partial_u.
\end{align*}

The prolongations of these evolutionary representatives acting on the PDE $\Delta_B$ are:
\begin{align}\label{eq:burgers_sym}
& \operatorname{pr} \left[v_1\right]_Q \left[\Delta_B\right] = -D_x \left[\Delta_B\right], \\ \nonumber
& \operatorname{pr} \left[v_2\right]_Q \left[\Delta_B\right] = -D_t \left[\Delta_B\right], \\ \nonumber
& \operatorname{pr} \left[v_3\right]_Q \left[\Delta_B\right] = -(3 + x D_x + 2 t D_t) \left[\Delta_B\right], \\ \nonumber
& \operatorname{pr} \left[v_4\right]_Q \left[\Delta_B\right] = -t D_x \left[\Delta_B\right], \\ \nonumber
& \operatorname{pr} \left[v_5\right]_Q \left[\Delta_B\right] = -t (3 + x D_x + t D_t) \left[\Delta_B\right]. \nonumber
\end{align}

However if one were to use the method of \citep{akhoundsadegh2023lie,zhang2023enforcing}, the resulting actions would instead be:
\begin{align}\label{eq:burgers_sym_tara}
& \operatorname{pr} v_1 \left[\Delta_B\right] = 0, \\ \nonumber
& \operatorname{pr} v_2 \left[\Delta_B\right] = 0, \\ \nonumber
& \operatorname{pr} v_3 \left[\Delta_B\right] = 3\nu\Delta_B, \\ \nonumber
& \operatorname{pr} v_4 \left[\Delta_B\right] = 0, \\ \nonumber
& \operatorname{pr} v_5 \left[\Delta_B\right] = 3\nu t\Delta_B. \nonumber
\end{align}

These calculations show that using the evolutionary representatives of the symmetries results in useful training signals, unlike previous methods used for loss augmentation.

\section{Darcy Flow Symmetries}
\label{ap:darcy}

In this appendix, we provide detailed derivations of the evolutionary representatives and their prolongations for the Darcy flow symmetries.

The Darcy flow equation is given for $x \in D \subset \mathbb{R}^2$:
\[
\Delta_D \coloneqq \nabla \cdot \left(k(x) \nabla u(x)\right) + f(x) = 0, 
\]
Unlike the the Burgers case, the algebra for the Darcy Flow is no longer finite dimensional. The algebra of symmetries for the Darcy flow equation includes the following generators:
\begin{align*}
& v^{\infty}_1 = h^{[1]}(u) \partial_u - k h^{[1]}_u(u) \partial_k, \\
& v^{\infty}_2 = -h^{[2]}_y(x, y) \partial_x - h^{[2]}_x(x, y) \partial_y + 2 f h^{[2]}_{xy}(x, y) \partial_f,
\end{align*}
where $h^{[1]}$ is arbitrary and $\nabla^2 h^{[2]}(x, y) = 0$.

The evolutionary representatives of these generators are:
\begin{align*} 
& v^{\infty}_1 = h^{[1]}(u)\partial_u - k h^{[1]}_u(u) \partial_k,  \\ 
& v^{\infty}_2=-h^{[2]}_y(x,y) \partial_x  -h^{[2]}_x(x,y) \partial_y \\ & \qquad\qquad\qquad  + 2fh^{[2]}_{xy}(x,y)\partial_f,
\end{align*}

The prolongations of these evolutionary representatives acting on the PDE are:
\begin{align*}
& \operatorname{pr} \left[v^{\infty}_1\right]_Q \left[\Delta_D\right] = 0, \\
& \operatorname{pr} \left[v^{\infty}_2\right]_Q \left[\Delta_D\right] = D_x [h^{[2]}_y \Delta_D] + D_y [h^{[2]}_x \Delta_D].
\end{align*}
Unfortunately, by the virtue of already being in its evolutionary form, the first vector does not result in a useful training signal. 

Now, if we are to enforce this during training we need to somehow sample $h^{[2]}$, such that it solves the Laplace equation. For practical purposes, we consider a finite-dimensional sub-algebra with linear $h^{[2]}$, resulting in a two-dimensional Lie algebra:
\begin{align}\label{eq:darcy_sym}
& \operatorname{pr} \left[v^{0}_2\right]_Q \left[\Delta_D\right] = D_y [\Delta_D], \\ \nonumber
& \operatorname{pr} \left[v^{1}_2\right]_Q \left[\Delta_D\right] = D_x [\Delta_D]. \nonumber
\end{align}

However if one were to use the method of \citep{akhoundsadegh2023lie,zhang2023enforcing}, the resulting actions would instead be:
\begin{align}\label{eq:darcy_sym_tara}
& \operatorname{pr} v^{\infty}_1 \left[\Delta_D\right] = 0, \\ \nonumber
& \operatorname{pr} v^{\infty}_2 \left[\Delta_D\right] = 2h^{[2]}_{xy}\Delta_D, \nonumber
\end{align}
which for the two dimensional sub-algebra would both be zero, resulting in no training signal. Thus, one would be forced to solve the Laplace equation, making the resulting method significantly more complex. 
\end{multicols}

\end{document}